\documentclass[10pt,journal,compsoc]{IEEEtran}

\usepackage{blindtext}
\usepackage[utf8]{inputenc} 
\usepackage[T1]{fontenc}    
\usepackage[colorlinks=true]{hyperref}       
\usepackage{url}            
\usepackage{booktabs}       
\usepackage{amsfonts}       
\usepackage{nicefrac}       
\usepackage{microtype}      
\usepackage{graphicx}
\usepackage{float}
\usepackage{sidecap}
\usepackage{amsmath}
\usepackage{accents}
\usepackage{todonotes}
\usepackage{subfigure}
\usepackage{float}
\usepackage{array}
\usepackage{multirow}
\usepackage{arydshln}
\usepackage[T1]{fontenc}
\usepackage[utf8]{inputenc}
\usepackage{tabularx,ragged2e,booktabs,caption}
\newcolumntype{C}[1]{>{\Centering}m{#1}}

\newcolumntype{K}{>{\arraybackslash}m{4cm}}
\newcolumntype{M}{>{\centering\arraybackslash}m{1cm}}
\newcolumntype{f}{>{\centering\arraybackslash}m{0.2cm}}

\begin{document}


\title{Moments in Time Dataset: one million \\ videos for event understanding} 

%

\author{Mathew Monfort, Alex Andonian, Bolei Zhou, 
\\Kandan Ramakrishnan, Sarah Adel Bargal, Tom Yan, Lisa Brown,
\\Quanfu Fan, Dan Gutfruend, Carl Vondrick, Aude Oliva%
\thanks{M Monfort, A Andonian, K Ramakrishnan, T Yan, A Oliva are with
Massachusetts Institute of Technology, 77 Massachusetts Ave, Cambridge, MA 02139 USA.}
\thanks{L Brown, Q Fan, D Gutfreund are with
International Business Machines, 75 Binney St., Cambridge, MA 02142 USA.}%
\thanks{B Zhou is with The Chinese
University of Hong Kong, Pond Cres, Ma Liu Shui, Hong Kong PRC}%
\thanks{C Vondrick is with Columbia University, 530 West 120th St, New York, NY 10027 USA.}%
\thanks{SA Bargal is with Boston University, 111 Cummington Mall, Boston, MA 02215 USA.}
}

\IEEEtitleabstractindextext{%
\begin{abstract}
We present the Moments in Time Dataset, a large-scale human-annotated collection of one million short videos corresponding to dynamic events unfolding within three seconds. Modeling the spatial-audio-temporal dynamics even for actions occurring in 3 second videos poses many challenges: meaningful events do not include only people, but also objects, animals, and natural phenomena;  visual and auditory events can be symmetrical in time ("opening" is "closing" in reverse), and either transient or sustained. We describe the annotation process of our dataset (each video is tagged with one action or activity label among 339 different classes), analyze its scale and diversity in comparison to other large-scale video datasets for action recognition, and report results of several baseline models addressing separately, and jointly, three modalities: spatial, temporal and auditory. The Moments in Time dataset, designed to have a large coverage and diversity of events in both visual and auditory modalities, can serve as a new challenge to develop models that scale to the level of complexity and abstract reasoning that a human processes on a daily basis.
\end{abstract}
\begin{IEEEkeywords}
video dataset, action recognition, event recognition
\end{IEEEkeywords}
}

\maketitle

\IEEEraisesectionheading{\section{Introduction}}

``The best things in life are not things, they are moments'' of raining, walking, splashing, 
jumping, etc. Moments happening in the world unfold at time scales from a second to minutes, occur in different places, and involve people, animals, objects and natural phenomena. 
Of particular interest are moments of a few seconds as they represent an ecosystem 
of diverse visual and auditory dynamic events.

We introduce the Moments in Time Dataset, a collection of one million short videos each with a label corresponding to 
an event unfolding in 3 seconds.\footnotemark\ 
Temporal events of such length correspond to the average duration of human working memory \cite{Baddeley1992,Barrouillet2004} which is a short-term memory-in-action buffer specialized in representing information that is changing over time.
Additionally, three seconds is a temporal envelope which holds meaningful actions between people, objects and phenomena (e.g. wind blowing, objects falling on the floor, 
shaking hands, playing with a pet, etc).


Compound activities that occur at longer time scales can be represented by sequences of three second actions.  For example, picking up an object and running could be interpreted as the compound actions "stealing", "saving" or  "playing sports" depending on the context of the activity (e.g. agent and scene). Hypothetically, when describing such a "stealing" event, one can go into the details of the movement of each joint and limb of the persons involved. However, this is not how we naturally describe compound events. Instead, we use verbs such as "picking" and "running" which are the actions which typically occur in a time window of 1-3 seconds. The ability to automatically recognize these short actions is a core step for automatic video comprehension.


\begin{figure*}
\includegraphics[width=\linewidth]{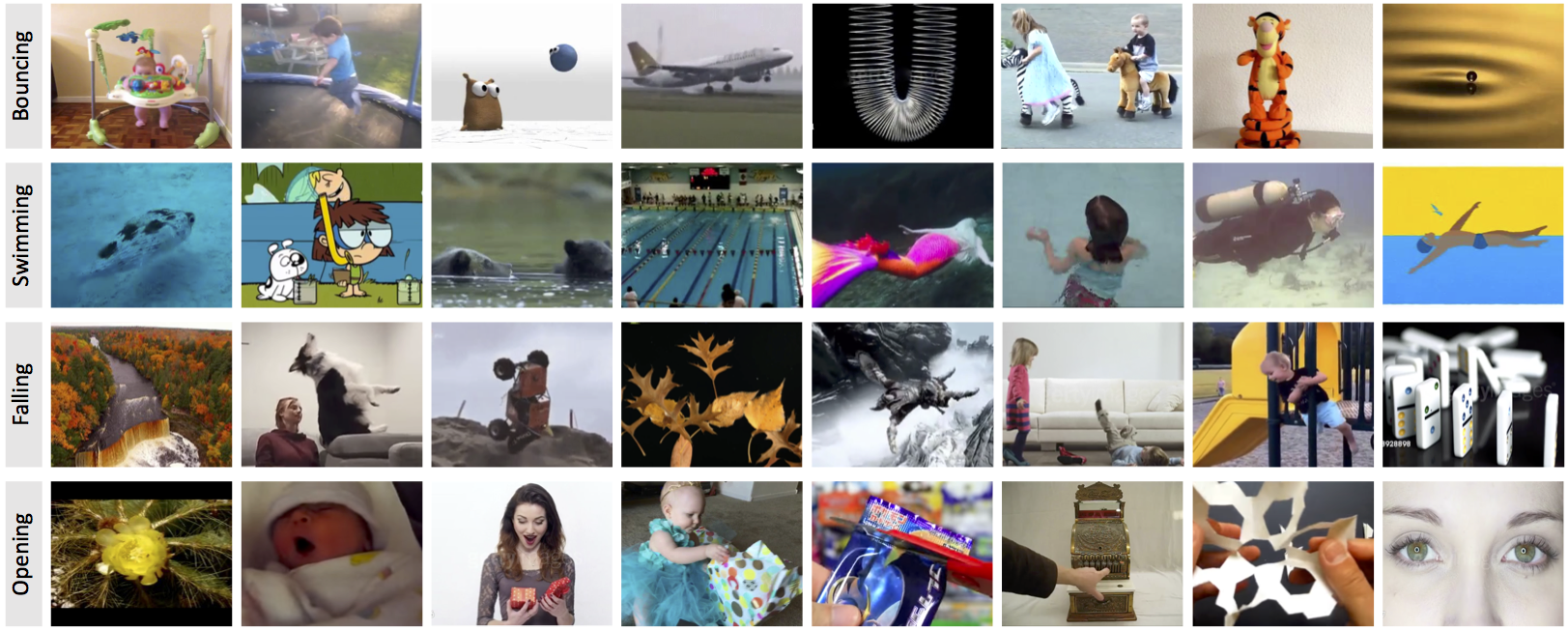}
\vspace{-2em}
\caption{\textbf{Sample Videos.} Day-to-day events can happen to many types of actors, in different environments, and at different scales. Moments in Time dataset has a significant intra-class variation among the categories. Here we illustrate one frame for a few video samples and actions. For example, car engines, books, and tulips can all open.}
\label{fig:diversity}
\end{figure*}

\footnotetext{The website is \url{http://moments.csail.mit.edu}}

Modeling the spatial-temporal dynamics even for three second videos, poses a daunting challenge. For instance, 
videos with the action "opening" include people opening doors, gates, drawers, and curtains, 
animals and humans opening eyes, 
and even a flower opening its petals. In some cases the same set of frames in reverse can actually depict a different action ("closing") showing that the temporal aspect is crucial to video understanding. Humans can recognize a common transformation that occurs in space and time that allows for all of the mentioned scenarios to be assigned to the category "opening" even though visually they look very different from each other. The challenge is to develop models that recognize these transformations in a way that will allow them to discriminate between different actions, yet generalize to other agents and settings within the same class.

We present the Moments in Time dataset, one million videos each with one action label from 339 different classes, to enable models to richly understand actions and dynamics in videos. This is one of the largest human-annotated video datasets capturing visual and audible short events produced by humans, animals, objects or nature. The most commonly used verbs in the English language  are chosen as the vocabulary covering a wide and diverse semantic space.  This presents a uniquely challenging and diverse dataset for action recognition with significant intra-class variation.  In Section \ref{sec:experiments} we include baseline results of several known models trained and tested on the dataset, addressing separately, and jointly, three modalities: spatial, temporal and auditory.

\section{Related Work}

\textbf{Video Datasets:} 
Over the years, the size of datasets for video understanding has grown steadily. KTH \cite{schuldt2004recognizing} and Weizmann \cite{blank2005actions} were early datasets for human action understanding. 
UCF101 \cite{soomro2012ucf101} and THUMOS \cite{jiang2014thumos} are built from web videos and have become important benchmarks for video classification. 
Kinetics \cite{kay2017kinetics} and YouTube-8M \cite{abu2016youtube} introduced a large number of event classes by leveraging public videos from YouTube. The micro-videos dataset \cite{nguyen2016open} uses social media videos to study an open-world vocabulary for video understanding. ActivityNet \cite{caba2015activitynet} explores recognizing activities in video and AVA \cite{gu2017ava} explores recognizing amd localizing fine-grained actions. The ``something something'' dataset \cite{goyal2017something} and Charades \cite{DBLP:journals/corr/SigurdssonVWFLG16} used crowdsourced workers to collect video datasets while the VLOG dataset \cite{fouhey2017lifestyle} 
collects daily human activities with natural spatio-temporal context.

\textbf{Video Classification:} The availability of large-scale video datasets has enabled significant progress in video understanding and classification. In early work, Laptev and Lindeberg \cite{laptev2003space} developed space-time interest point descriptors and Klaser et al.\ \cite{klaser2008spatio} designed histogram features for video. Pioneering work by Wang et al.\ \cite{wang2011action} developed dense action trajectories by separating foreground motion from camera motion. Sadanand and Corso \cite{sadanand2012action} designed ActionBank as a high-level representation for video and action classification, and Pirsiavash and Ramanan \cite{pirsiavash2014parsing} leveraged grammar models for temporally segmenting actions from video. Advances in deep convolutional networks have enabled the development of a variety of large-scale video classification models. Various approaches of fusing RGB frames over the temporal dimension are explored on the Sport1M dataset \cite{karpathy2014large}. Two stream CNNs with one static image stream and one optical flow stream were proposed to fuse the information of object appearance and short-term motion \cite{simonyan2014two}. 3D convolutional networks \cite{tran2015learning} use 3D kernels to extract features from a sequence of RGB frames. Temporal Segment Networks sample frames and optical flow on different time scales to extract information for activity recognition \cite{wang2016temporal}. A CNN+LSTM model, which uses a CNN to extract frame features and an LSTM to integrate features over time, is also used to recognize activities in videos \cite{donahue2015long}. Recently, I3D networks \cite{carreira2017quo} use two stream CNNs with inflated 3D convolutions on both RGB and optical flow sequences to achieve state of the art results on the Kinetics dataset \cite{kay2017kinetics}.   More recent 3D networks incorporate Non-local modules \cite{Wang_2018_CVPR} in order to capture long-range dependencies while Temporal Relation Networks \cite{zhou2017temporal} take a different approach by learning the relevant state transitions in sparsely sampled frames from different temporal segments.



\textbf{Sound Classification:} Environmental and ambient sound recognition is a rapidly growing area of research. Stowell et al.\ \cite{stowell2015detection} collected an early dataset and assembled a challenge for sound classification, Piczak \cite{piczak2015environmental} collected a dataset of fifty sound categories and enough to train deep convolutional models,  Salamon et al.\ \cite{salamon2014dataset} released a dataset of urban sounds, and Gemmeke et al.\ \cite{gemmeke2017audio} use web videos for sound dataset collection. Recent work is now developing models for sound classification with deep neural networks. For example, Piczack \cite{piczak2015environmental} pioneered early work for convolutional networks for sound classification, Aytar et al.\  \cite{NIPS2016_6146} transfer visual models into sound for auditory analysis, and Hershey et al.\ \cite{hershey2016cnn} develop large-scale convolutional models for sound classification, and Arandjelovi{\'c} and Zisserman \cite{arandjelovic2017look} train sound and vision representations jointly. In Moments in Time dataset, many videos have both visual and auditory signals, enabling for multi-modal video recognition. 



\section{The Moments in Time Dataset}

The goal of this project is to design a high-coverage, high-density, balanced dataset of hundreds of verbs depicting moments of a few seconds. High-quality datasets should have a broad coverage, high diversity and density of samples, and the ability to scale.  The Moments in Time Dataset consists of over one million 3-second videos corresponding to 339 different verbs. Each verb is associated with over 1,000 videos resulting in a large balanced dataset for learning dynamic events from videos. 
Importantly, the dataset is designed to have, and grow towards, a very large set of both inter-class and intra-class variation that captures a dynamic event at different levels of abstraction (i.e. "opening" doors, curtains, eyes, mouths, even a flower opening its petals).

\subsection{Building a Vocabulary of Active Moments}
\label{subsec:vocabulary}
We began building our vocabulary by forming a list of the 4,500 most commonly used verbs from VerbNet \cite{verbNet} (according to the word frequencies in the Corpus of Contemporary American English (COCA) \cite{coca}).  We clustered these verbs using features from Propbank \cite{propBank}, FrameNet \cite{frameNet} and OntoNotes \cite{ontonotes} which contain information on both the meaning and usage of each verb.  This allows for clusters to be formed that extend beyond synonymous groups.  

We formed our clusted by assigning a binary feature vector to each verb with an index for different features provided by FrameNet, PropBank and VerbNet.  If a verb was associated with a given feature it's value was set to 1 and 0 if not.  These feature vectors were then used to perform k-means clustering to create the set of semantic verb clusters.  For example, the actions "pour" and "flow" have similar meanings and can both be used to describe the movement of a liquid.  However, "pour" has features associated with an agent (e.g. a person pouring coffee) whereas "flow" does not.  These differences can lead to very different video clips when we take into account the full context of an event.

To order our list of verbs we iteratively selected the most common verb from the cluster with the highest cumulative frequency of use among its members and added it to our vocabulary.  For example, a cluster associated with "grooming" contains the following verbs in order of most common to least common "washing, showering, bathing, soaping, grooming, shampooing, manicuring, moisturizing, and flossing".  Verbs can belong to multiple clusters due to their different frames of use.  For instance, "washing" also belongs to a group associated with cleaning, mopping, scrubbing, etc.  Once a verb was chosen it was then removed from all of its member clusters.  We repeated this process for all verbs in the set excluding verbs that were either ambiguous, not likely to be visual/audible in a 3-seconds video (e.g. "thinking" and "being") or too similar to a previously selected verb. We settled on a set of 339 frequently used and semantically diverse verbs that we used to build the proposed dataset with a large coverage and diversity of labels.

\subsection{Collection and Annotation}

To generate candidate videos for annotation, we search the Internet by parsing video metadata and crawling search engines to build a list of candidate videos for each class in our vocabulary using a variety of different sources \footnote{Youtube, Flickr, Vine, Metacafe, Peeks, Vimeo, VideoBlocks, Bing, Giphy, The Weather Channel, and Getty-Images}.  We download each video and randomly cut a 3-second section which we group with the corresponding verb. These verb-video tuples are then sent to Amazon Mechanical Turk (AMT) for annotation. To ensure the highest level of diversity, we cut a single 3-second snippet from each video source.  We used this approach instead of using a model to localize interesting video segments, such as Video2Gif \cite{Gygli2016Video2GIFAG}, to reduce any model bias where a trained model will localize segments more similar to the data for which it was trained.

\begin{figure}[h]
  \centering
  \fbox{\includegraphics[width=0.5\linewidth, trim={7cm 0.4cm 6cm 0.8cm},clip]{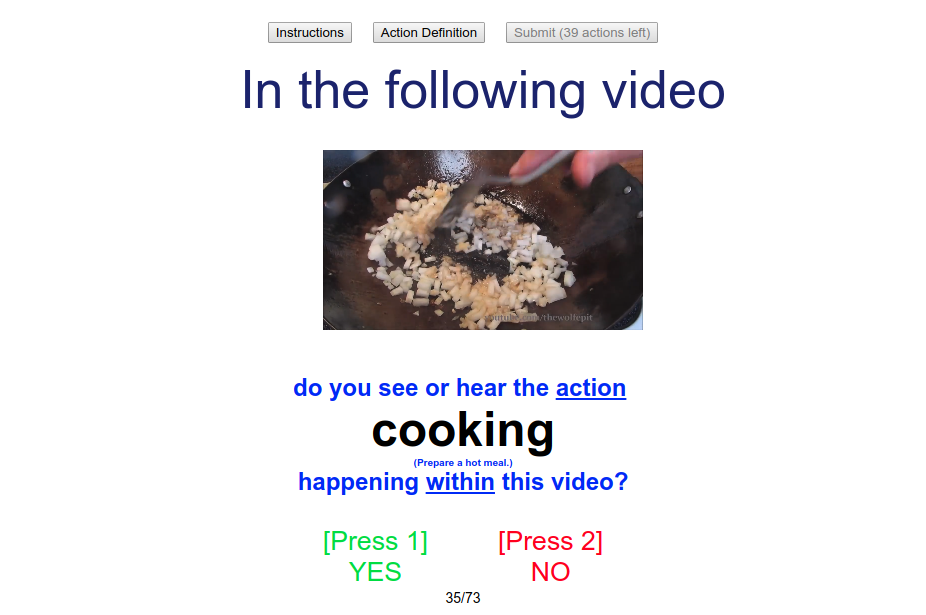}}
  \caption{\textbf{User interface.} An example for our binary annotation task for the action cooking.}
  \label{fig:annotationExample}
\end{figure}

Each AMT worker is presented with a video-verb pair and asked to press a Yes or No key signifying if the action is happening in the scene.  Positive responses from the first round are sent to subsequent rounds of annotation.  Each HIT (a single worker assignment) contains 64 different 3-second videos that are related to a single verb and 10 ground truth videos that are used for control. In each HIT, the first 4 questions are used to train the workers on the task and require the correct answer to be selected before continuing. Only the results from HITs that earn a 90\% or above on the control videos are included in the dataset.  This binary-classification setup eases class selection for workers allowing for efficient annotation. We run each video in the training set through annotation at least 3 times and require a human consensus of at least 75\% to be considered a positive label.  For the validation and test set we increase the minimum number of rounds of annotation to 4 with a human consensus of at least 85\%.  We do not set the threshold at 100\% to allow for  videos with more difficult to recognize actions. Figure \ref{fig:annotationExample} shows an example of the annotation task. 


\subsection{Dataset Statistics}

\begin{figure*}
	\centering
  \begin{minipage}[h]{0.33\linewidth}
  	\centering
    \includegraphics[width=\linewidth]{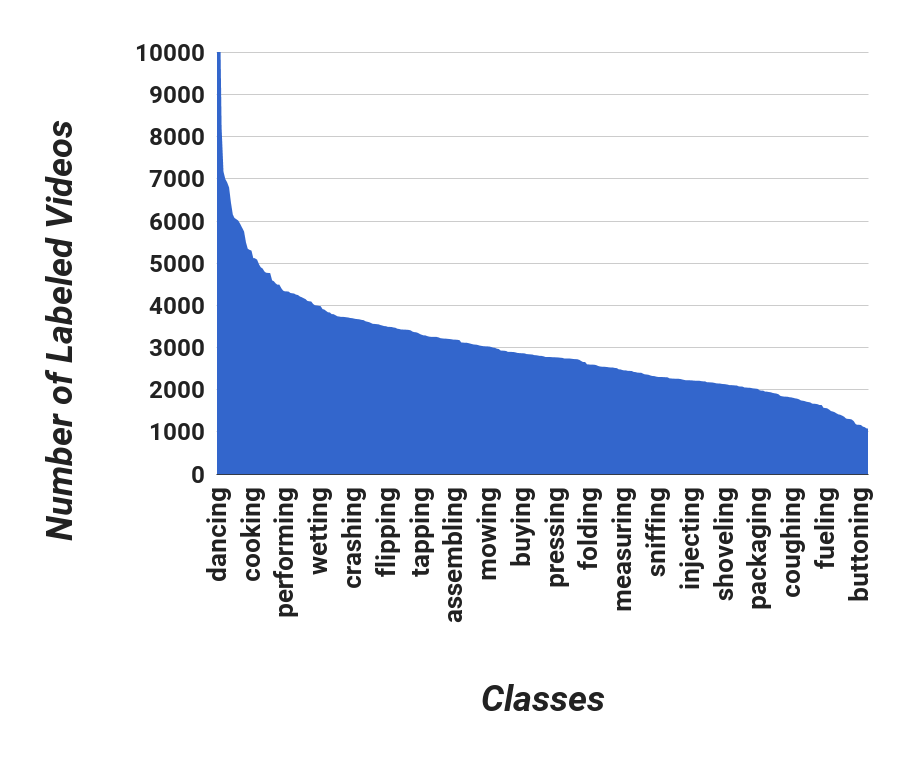}
  \end{minipage}
  \begin{minipage}[h]{0.33\linewidth}
  	\centering
    \includegraphics[width=\linewidth]{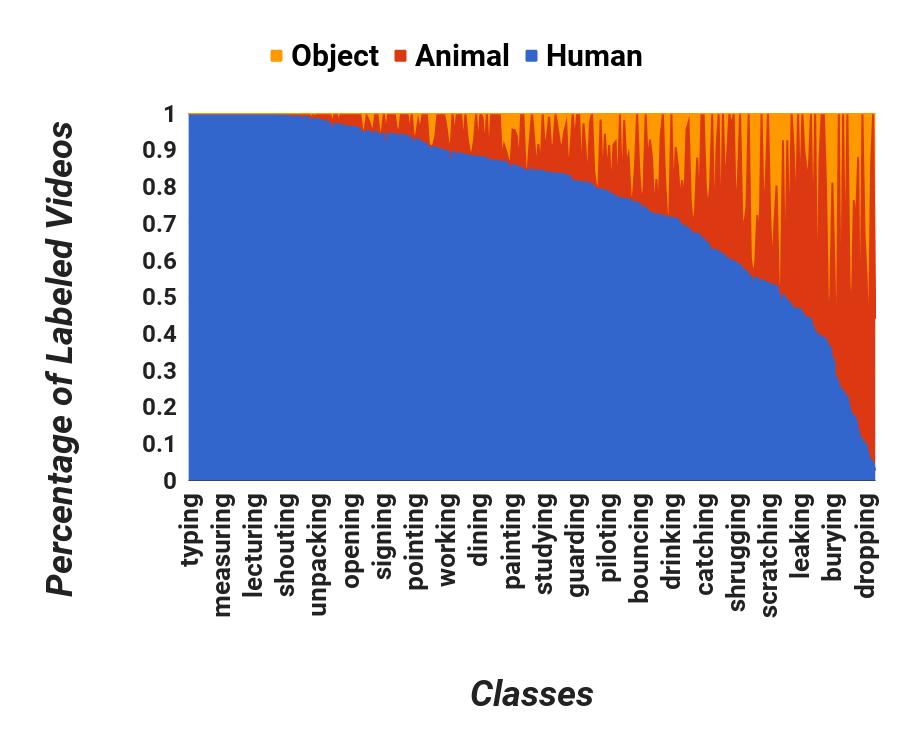}
  \end{minipage}
  \begin{minipage}[h]{0.33\linewidth}
  	\centering
    \includegraphics[width=\linewidth]{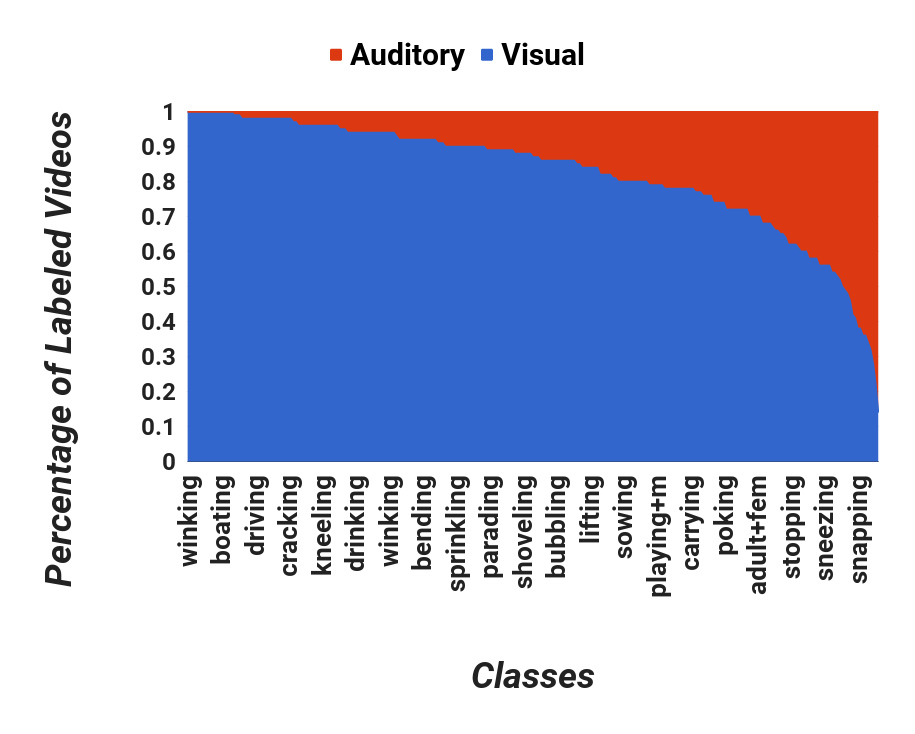}
  \end{minipage}
  \vspace{-1em}
    \caption{\textbf{Dataset Statistics.} {\bf{Left}}: Distribution of the number of videos belonging to each category. {\bf{Middle}}: Per class distribution of videos that have humans, animals, or objects as agents completing actions. {\bf{Right}}: Per class distribution of videos that require audio to recognize the class category and videos that can be categorized with only visual information.}
    \label{fig:datasetStats}
\end{figure*}
\begin{figure*}
	\centering
  \begin{minipage}[h]{0.33\linewidth}
  	\centering
    \includegraphics[width=\linewidth]{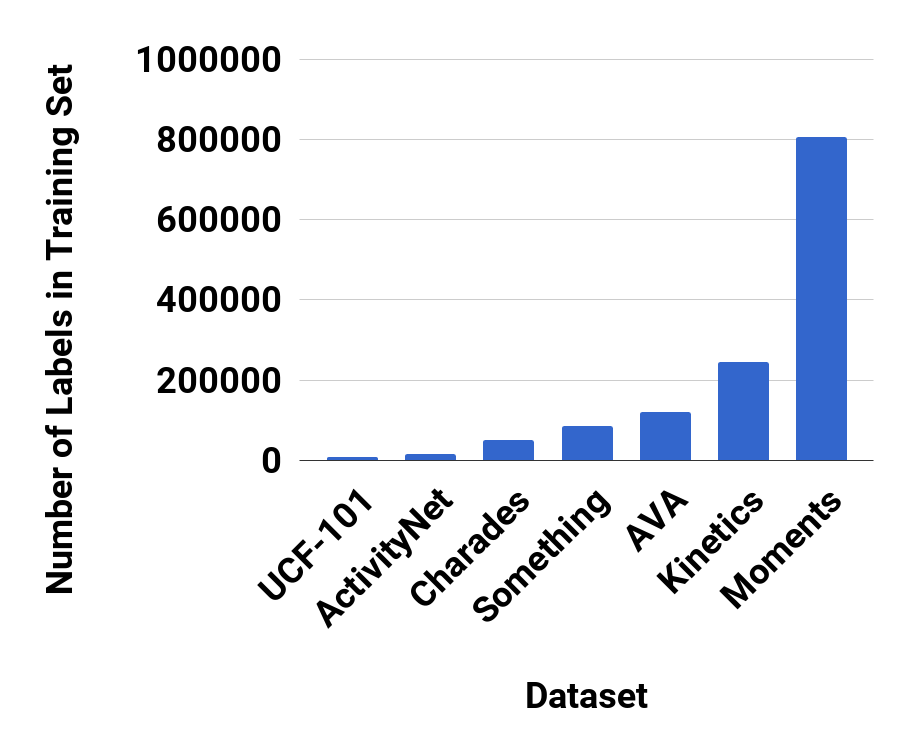}
  \end{minipage}
  \begin{minipage}[h]{0.33\linewidth}
  	\centering
    \includegraphics[width=\linewidth]{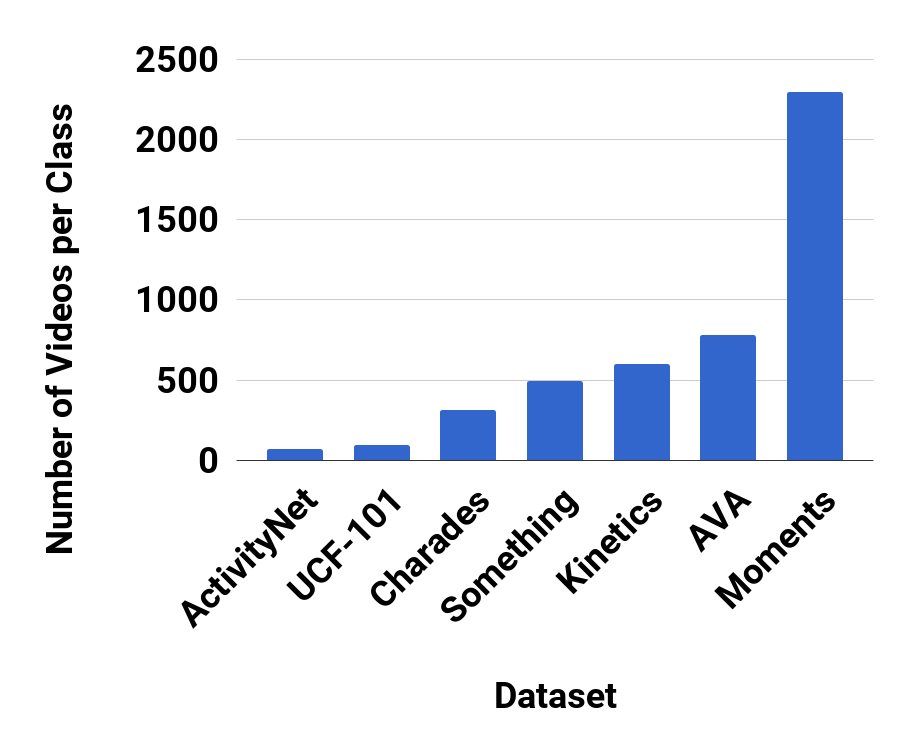}
  \end{minipage}
  \begin{minipage}[h]{0.33\linewidth}
  	\centering
    \includegraphics[width=\linewidth]{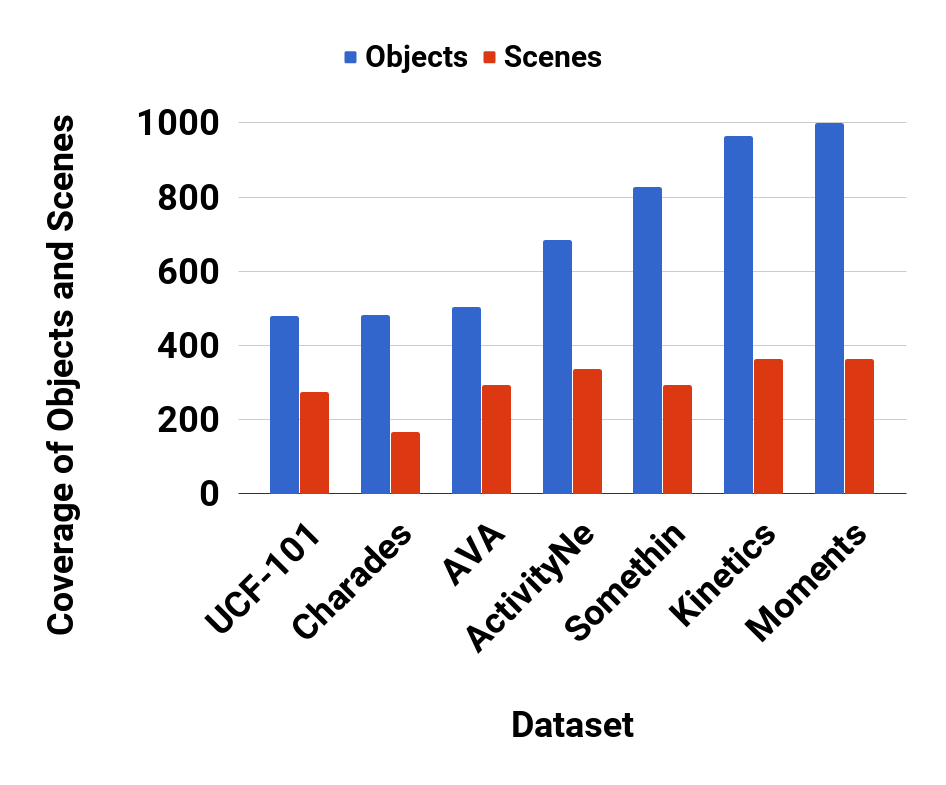}
  \end{minipage}
    \vspace{-1em}
    \caption{\textbf{Comparison to Datasets.} For each dataset we provide different comparisons. {\bf{Left}}: the total number of action labels in the training set. {\bf{Middle}}: the average number of videos per class (some videos can belong to multiple classes).{\bf{Right}}: the coverage of objects and scenes recognized (top 1) by networks trained on Places and Imagenet.}
    \label{fig:datasetComparison}
\end{figure*}

A motivation for this project was to gather a large balanced and diverse dataset for training models for video understanding.  Since we pull our videos from over 10 different sources we are able to include a large breadth of diversity that would be challenging using a single source.
In total, we have collected over 1,000,000 labelled videos for 339 Moment classes. The graph on the left of Figure \ref{fig:datasetStats} shows the full distribution across all classes where the average number of labeled videos per class is 1,757 with a median of 2,775.
 
To further aid in building a diverse dataset we do not restrict the active agent in our videos to humans.  Many events such as "walking", "swimming", "jumping", and "carrying" are not specific to human agents.  In addition, some classes may contain very few videos with human agents (e.g. "howling" or "flying").  True video understanding models should be able to recognize the event across agent classes.  With this in mind we decided to build our dataset to be general across agents and present a new challenge to the field of video understanding.  The middle graph in Figure \ref{fig:datasetStats} shows the distribution of the videos according to agent type (human, animal, object) for each class.  On the far left (larger human proportion), we have classes such as "typing", "sketching", and "repairing", while on the far right (smaller human proportion) we have events such as "storming", "roaring", and "erupting".

Another feature of the Moments in Time dataset is that we include sound-dependant classes.  We do not restrict our videos to events that can be seen, if there is a moment that can only be heard in the video (e.g. "clapping" in the background) then we still include it.  This presents another challenge in that purely visual models will not be sufficient to completely solve the dataset.  The right graph in Figure \ref{fig:datasetStats} shows the distribution of videos according to whether or not the event in the video can be seen.

\subsection{Dataset Comparisons}

In order to highlight the key points of our dataset, we compare the scale and object-scene coverage found in Moments in Time to other large-scale video datasets for action recognition.  These include UCF101 \cite{soomro2012ucf101}, ActivityNet \cite{caba2015activitynet}, Kinetics \cite{kay2017kinetics}, Something-Something \cite{goyal2017something}, AVA \cite{gu2017ava}, and Charades \cite{DBLP:journals/corr/SigurdssonVWFLG16}.
Figure \ref{fig:datasetComparison} compares the total number of action labels used for training (left) and the average number of videos that belong to each class in the training set (middle).  This increase in scale  for action recognition is beneficial for training large generalizable systems for machine learning.

Additionally, we compared the coverage of objects and scenes that can be recognized within the videos.  This type of comparison helps to showcase the visual diversity of our dataset.  To accomplish this, we extract 3 frames from each video evenly spaced at 25\%, 50\%, and 75\% of the video duration and run a 50 layer resnet \cite{resNet} trained on ImageNet \cite{NIPS2012_4824} and a 50 layer resnet trained on Places \cite{zhouKLTO16} over each frame and average the prediction results for each video.  We then compare the total number of objects and scenes recognized (top 1) by the networks in Figure \ref{fig:datasetComparison} (right).  The graph shows that 100\% of the scene categories in Places and 99.9\% of the object categories in ImageNet were recognized in our dataset.  The closest dataset to ours in this comparison is Kinetics which has a recognized coverage of 99.5\% of the scene categories in Places and 96.6\% of the object categories in ImageNet.  We should note that we are comparing the recognized categories from the top 1 prediction of each network.  We have not annotated the scene locations and objects in each video of each dataset.  However, a comparison of the visual features recognized by each network does still serve as an informative comparison of visual diversity.

\section{Experiments}
\label{sec:experiments}

In this section we present the details of our experimental setup utilized to obtain the reported baseline results. 

\subsection{Experimental Setup}

\textbf{Data.} We generate a training set of 802,264 videos with between 500 and 5,000 videos per class for 339 different classes and evaluate performance on a validation set of 33,900 videos with 100 videos for each class.  We additionally withhold a test set of 67,800 videos consisting of 200 videos per class which will be used to evaluate submissions for a future action recognition challenge.  

\textbf{Preprocessing.} We extract RGB frames from the videos at 25 fps and resize the RGB frames to a standard 340x256 pixels. In the interest of performance, we pre-compute optical flow on consecutive frames using an off-the-shelf implementation of TVL1 optical flow algorithm \cite{zach2007duality} from the OpenCV toolbox \cite{itseez2015opencv}. This formulation allows for discontinuities in the optical flow field and is thus more robust to noise. For fast computation, we discretize the values of optical flow fields into integers, clip the displacement with a maximum absolute value of 15 and scale the range to 0-255. The x and y displacement fields of every optical flow frame are then stored as two grayscale images to reduce storage. To correct for camera motion, we subtract the mean vector from each displacement field in the stack. For video frames, we use random cropping for data augmentation and subtract the ImageNet mean from images. 

\begin{figure*}[tb]
  	\centering
    \includegraphics[width=\linewidth]{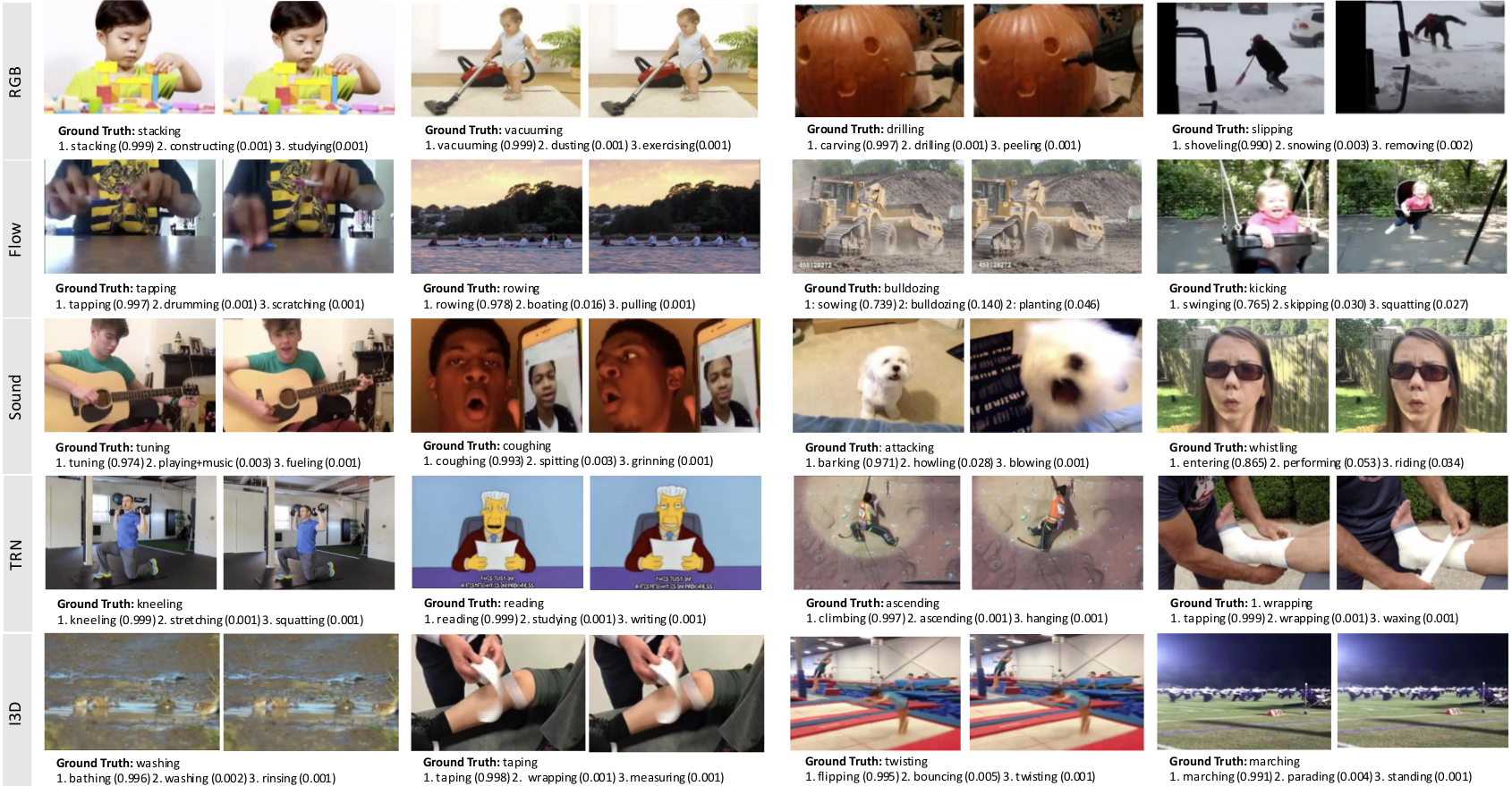}
    \caption{\textbf{Overview of top detections for several single stream models.} The ground truth label and top three model predictions are listed for representative frames of videos.}
    \label{prediction_result}
\end{figure*}

\textbf{Evaluation metric.} We use top-1 and top-5 classification accuracy as the scoring metrics. Top-1 accuracy indicates the percentage of testing videos for which the top confident predicted label is correct. Top-5 accuracy indicates the percentage of the testing videos for which the ground-truth label is among the top 5 ranked predicted labels. This is appropriate for video classification as videos may contain multiple actions (see Figure \ref{prediction_result}). For evaluation we randomly select 10 crops per frame and average the results.

\begin{table}[tb]
\footnotesize
\centering
\begin{tabular}{ l  c  c  r }
\toprule      \textbf{Model} & \textbf{Modality} & \textbf{Top-1 (\%)}  & \textbf{Top-5 (\%)} \\ 
\hline
 Chance &- & 0.29 & 1.47 \\
\hline  
ResNet50-scratch & Spatial & 23.65 & 46.73 \\
ResNet50-Places & Spatial & 26.44 & 50.56 \\
ResNet50-ImageNet & Spatial & 27.16 & 51.68 \\
TSN-Spatial & Spatial & 24.11 & 49.10 \\
\hline
BNInception-Flow & Temporal & 11.60 & 27.40 \\
TSN-Flow & Temporal & 15.71 & 34.65 \\
\hline
SoundNet & Auditory & 7.60  & 18.00 \\
\hline
TSN-2stream & Spatial+Temporal & 25.32 & 50.10 \\
TRN-Multiscale & Spatial+Temporal & 28.27 & 53.87 \\
I3D & Spatial+Temporal & 29.51 & 56.06 \\
\hline
Ensemble (SVM) & S+T+A & 31.16 & 57.67 \\ 
\bottomrule
\end{tabular}
\caption{\textbf{Classification Accuracy:} We show Top-1 and Top-5 accuracy of the baseline models on the validation set.}
\label{table:baselines}
\end{table}


\begin{figure*}[tb]
\centering
\includegraphics[width=1\linewidth]{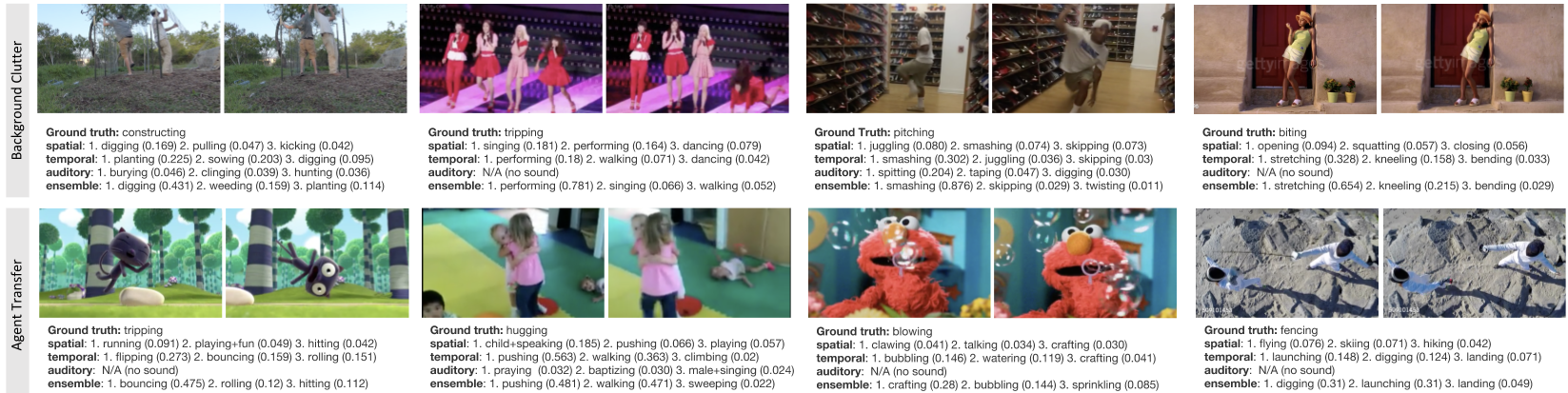}
\vspace{-1em}
\caption{\textbf{Examples of missed detections:} We show examples of videos where the prediction is not in the top-5. Common failures are often due to background clutter or poor generalization across agents (humans, animals, objects).}
\label{fig:errors}
\end{figure*}

\subsection{Baselines for Video Classification}
\label{sec:baselineModels}

\begin{figure*}
\centering
\includegraphics[width=1\linewidth]{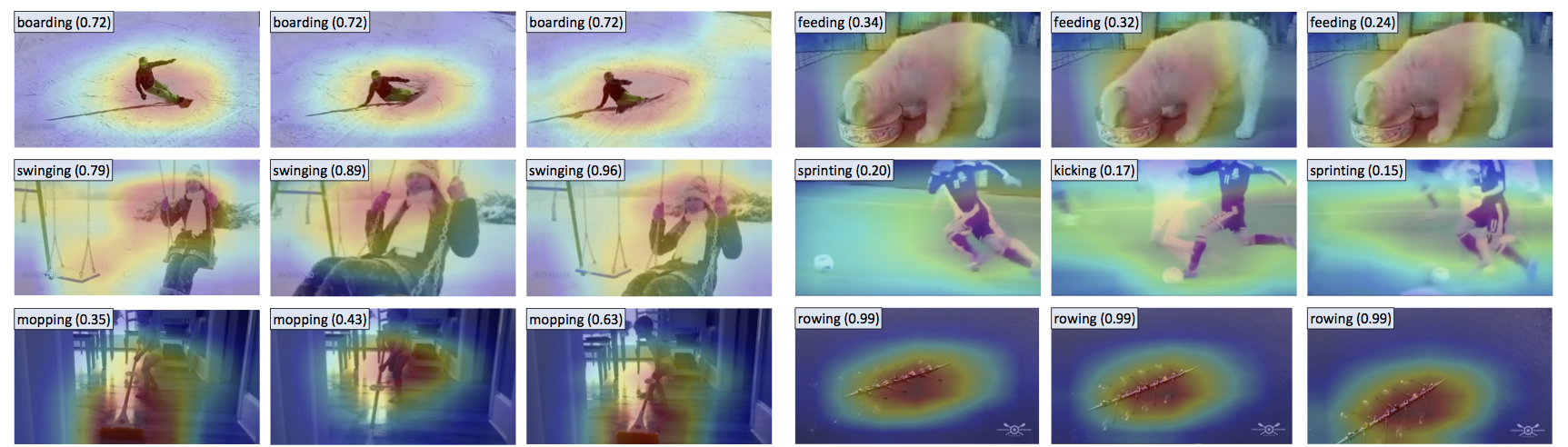}
\vspace{-2em}
\caption{\textbf{Predictions and Attention:} We show some predictions (shown with class probability in top left corner) from \textit{ResNet50-ImageNet} spatial model on held-out video data and the heatmaps which highlight the informative regions in some frames. For example, for recognizing the action chewing, the network focuses on the moving mouth.}
\label{fig:attention}
\end{figure*}

Here, we present several baselines for video classification on the Moments in Time dataset. We show results for three modalities (spatial, temporal, and auditory), as well as for recent video classification models such as Temporal Segment Networks \cite{wang2016temporal} and Temporal Relation Networks \cite{zhou2017temporal}. We further explore combining models to improve recognition accuracy. The details of the baseline models grouped by different modalities are listed below.

\textbf{Spatial modality}. We experiment with a 50 layer resnet (Resnet50) \cite{he2016identity} trained on randomly selected RGB frames from each video with networks trained from scratch (\textit{ResNet50-scrach}), initialized on Places \cite{zhou2014learning} (\textit{ResNet50-Places}), and initialized on ImageNet \cite{NIPS2012_4824} (\textit{ResNet50-ImageNet}). In testing, we average the prediction from 6 equi-distant frames.

\textbf{Auditory modality}. While many actions can be recognized visually, sound contains complementary or even mandatory information for recognition of particular classes, such as cheering or talking, as can be seen in Figure \ref{fig:datasetStats} (right). We use raw waveforms as the input modality and finetune a SoundNet network which was pretrained on 2 million unlabeled videos from Flickr \cite{NIPS2016_6146} with the output layer changed to predict moment classes (\textit{SoundNet}).

\textbf{Temporal modality}. Following \cite{simonyan2014two}, we compute the optical flow between adjacent frames encoded in Cartesian coordinates as displacements by stacking together 5 consecutive frames to form a 10 channel image (the x and y displacement channels).  We then modify the first convolutional layer of a BNInception \cite{ioffe2015batch} model to accept 10 input channels (\textit{BNInception-Flow}). 

\textbf{Spatial-Temporal modality}.
We also train three recent action recognition models: Temporal Segment Networks (TSN) \cite{wang2016temporal}, Temporal Relation Networks \cite{zhou2017temporal} and Inflated 3D convolutional networks (I3D) \cite{carreira2017quo}. Temporal Segment Networks aim to efficiently capture the long-range temporal structure of videos using a sparse frame-sampling strategy. The TSN's spatial stream \textit{TSN-Spatial} is fused with an optical flow stream \textit{TSN-Flow} via average consensus to form the two stream TSN \textit{TSN-2stream}. The base model for each stream is a BNInception \cite{ioffe2015batch} model with three time segments.

Temporal Relation Networks (TRN) \cite{zhou2017temporal} explicitly learn temporal dependencies between video segments that best characterize a particular action. This ``plug-and-play" module can simultaneously model several short and long range temporal dependencies to classify actions that unfold at multiple time scales. We trained a TRN with 8 multi-scale relations \textit{TRN-Multiscale} on RGB frames using a Resnet50 \cite{he2016identity} base model. Note that we classify the TRN-Multiscale as spatiotemporal modality because in training it utilizes the temporal dependency of different frames. 

Inflated 3D convolutional networks (I3D) \cite{carreira2017quo} \emph{inflate} the convolutional and pooling kernels of a pretrained 2D network to a third dimension. The inflated 3D kernel is initialized from the 2D model by repeating the weights from the 2D kernel over the temporal dimension. This improves learning efficiency and performance as 3D models contain far more parameters than their 2D counterpart and a strong intialization greatly improves training.  For our experiments we use a 3D Resnet50 inflated from our best 2D Resnet50 (spatial) from Table \ref{table:baselines}.  We train each model with 16 frames selected at 5 frames-per-second (fps) for each video.


\textbf{Ensemble}. To combine different modalities for action recognition, we form an ensemble using the top performing model of each modality (spatial: \textit{ResNet50-ImageNet}, spatiotemporal: \textit{I3D} and auditory: \textit{SoundNet}). We concatenate the features from the final hidden layer of each modality and train a linear SVM to predict the moment categories (SVM). This ensemble learns how to fuse the features of the different modalities in order to make a more robust prediction based on spatial, temporal and auditory information giving us our highest recognition scores of 31.16\% top-1 and 57.67\% top-5.

\begin{table}[tb]
\centering
\begin{tabular}{ c | c c c c}
      &  \multicolumn{3}{c}{\textbf{Fine-Tuned}} \\
       \textbf{Pretrained} & UCF & HMDB & Something \\ 
       \cmidrule{2-4}
      Kinetics & Top-1: 92.6  & Top-1: 62.0 & Top-1: 48.6  \\
        & Top-5: 99.2  & Top-5: 88.2 & Top-5: 77.9 \\
       \cmidrule{2-4}
       \cmidrule{2-4}
      Moments & Top-1: 91.9 & Top-1: 65.9 & Top-1: 50.0 \\
       & Top-5: 98.6 & Top-5: 89.3 & Top-5: 78.8\\
      \bottomrule
      \end{tabular}
    \caption{Dataset transfer performance using ResNet50 I3D models pretrained on both Kinetics and Moments in Time.}%
    \label{fig:cross_dataset_image}%
\end{table}

\subsection{Baseline Results}

Table \ref{table:baselines} shows the performance of the baseline models on the validation set. The best single model is I3D, with a Top-1 accuracy of 29.51\% and a Top-5 accuracy of 56.06\% while the Ensemble model (SVM) achieves a 57.67\% Top-5 accuracy.

Figure \ref{prediction_result} illustrates some of the high scoring predictions from the baseline models. These qualitative results suggest that the models can recognize moments well when the action is well-framed and close up. However, the model frequently misfires when the category is fine-grained or there is background clutter. Figure \ref{fig:errors} shows examples where the ground truth category is not detected in the top-5 predictions due to either significant background clutter or difficulty in recognizing actions across agents. 


We visualize the prediction given by the model by generating heatmaps for some video samples using Class Activation Mapping (CAM) \cite{zhou2016learning} in Figure \ref{fig:attention}. CAM highlights the most informative image regions relevant to the prediction. Here we use the top-1 prediction of the \textit{ResNet50-ImageNet} model for each individual frame of the given video.  

Categories that perform the best tend to have clear appearances and lower intra-class variation, for example bowling and surfing frequently happen in specific scene categories. The more difficult categories, such as covering, slipping, and plugging, tend to have wide spatiotemporal support as they can happen in most scenes and with most objects. Recognizing actions uncorrelated with scenes and objects seems to pose a challenge for video understanding.

Auditory models have qualitatively different performance per category versus visual models suggesting that sound provides a complementary signal to vision. However, the full ensemble model has per category performance that is fairly correlated with a single image (spatial) model. Given the relatively low performance on Moments in Time, this suggests that there is still room to capitalize on temporal and auditory dynamics to better recognize actions.


\subsection{Cross Dataset Transfer}

We additionally conducted a set of transfer experiments where we pretrain two models, one on Kinetics \cite{kay2017kinetics} and one on Moments in Time, in order to evaluate which model generalizes better to other datasets. We use a Resnet50 I3D model for the experiments as this model gave us the best single stream performance on our dataset (Table \ref{table:baselines}).  We compare our results when transferring to UCF101 \cite{soomro2012ucf101}, HMDB51 \cite{10.1007/978-3-642-33374-3_41} and Something-Something \cite{goyal2017something}.  During training we randomly crop the average duration of each video in each dataset at 5 frames-per-second (fps).  This allows for scalable transfer between datasets with different video lengths while keeping the frame-rates consistent with the dataset used for pretraining.  For evaluation we apply the model using a sliding window at 5 fps with a 2 frame step-size and average the results.


Table \ref{fig:cross_dataset_image} shows the results of the transfer task where the top-1 and top-5 scores are calculated by evaluating on the validation set of the dataset used to fine-tune the model.  We can see from the results that pretraining on Moments in Time results in better performance when transferring to HMDB51 and pretraining on Kinetics gives stronger results when transferring to UCF101.  This makes sense as UCF101 and Kinetics share many classes (e.g. "playing guitar", "skydiving", "walking dog", etc.) and both consist solely of YouTube videos, while HMDB51 is built from multiple sources and has a classes more similar to Moments in Time (e.g. "eating", "laughing", "pouring", etc.). The results on Something-Something show that pretraining on Moments in Time improves performance on a dataset designed for learning action concepts (e.g. picking \textit{something} up). Additionally, the fact that each dataset has videos which are consistently longer than 3 seconds (significantly so for HMDB51) suggests that the 3 second length of the videos in the Moments in Time dataset does not hinder performance when applied to datasets with much longer videos.

\section{Conclusion}

We present the Moments in Time Dataset, a large-scale collection of three second videos covering a wide range of dynamic events involving different agents (people, animals, objects, and natural phenomena). We report results of several baseline models addressing separately, and jointly, three modalities: spatial, temporal and auditory. This dataset presents a difficult task for the field of computer vision as the labels correspond to different levels of abstraction (a verb like "falling" can apply to many different agents and scenarios and involve objects and scenes of different categories, see Figure \ref{fig:diversity}). It will serve as a new challenge to develop models that can appropriately scale to the level of complexity and abstract reasoning needed to process each video.

\textbf{Acknowledgements:}
This work was supported by the MIT-IBM Watson AI Lab, the Intelligence Advanced Research Projects Activity (IARPA) via Department of Interior / Interior Business Center (DOI/IBC) contract number D17PC00341 and the Toyota Research Institute / MIT CSAIL Joint Research Center.

{
\bibliographystyle{plain}
\bibliography{main}
}

\end{document}